\documentclass{article}

\PassOptionsToPackage{numbers,compress}{natbib}
\usepackage[preprint]{neurips_2026}

\usepackage[utf8]{inputenc}
\usepackage[T1]{fontenc}
\usepackage{hyperref}
\usepackage{url}
\usepackage{booktabs}
\usepackage{amsfonts}
\usepackage{amsmath,amssymb}
\usepackage{nicefrac}
\usepackage{microtype}
\usepackage{xcolor}
\usepackage{array}
\usepackage{graphicx}
\usepackage{tikz}
\usepackage{pgfplots}

\pgfplotsset{compat=1.18}
\usetikzlibrary{arrows.meta,positioning,shapes.geometric,calc}

\hypersetup{
  colorlinks=true,
  linkcolor=black,
  citecolor=black,
  urlcolor=blue,
  pdfauthor={Anis Radianis},
  pdftitle={Learn-by-Wire Training Control Governance: Bounded Autonomous Training Under Stress for Stability and Efficiency}
}

\newcommand{\lbw}{LBW-Guard}
\newcommand{\adamw}{AdamW}

\title{Learn-by-Wire Training Control Governance:\\
Bounded Autonomous Training Under Stress for Stability and Efficiency}

\author{%
  Anis Radianis\\
  Research Scientist at Qluon Inc.\\
  Delaware, United States\\
  \texttt{aradianis@qluon.ai}\\
}
\begin{document}
\maketitle
\begin{center}
{\small \textit{Preprint version prepared for arXiv, May 2026.}}
\end{center}

\begin{abstract}
Modern language-model training is increasingly exposed to instability, degraded runs, and wasted compute, especially under aggressive learning-rate, scale, and runtime-stress conditions. This paper introduces Learn-by-Wire Guard (LBW-Guard), a bounded autonomous training-control governance layer that operates above AdamW. Rather than replacing the optimizer update rule, LBW-Guard observes training telemetry, interprets instability-sensitive regimes, and applies bounded control to optimizer execution while preserving fixed training objectives. 

We evaluate LBW-Guard in a Qwen2.5-centered stress-and-robustness suite using WikiText-103, with Qwen2.5-7B as the empirical anchor, model-size comparisons against Qwen2.5-3B and Qwen2.5-14B, learning-rate stress tests, gradient-clipping baselines, and a no-LoRA TinyLlama-1B full-parameter sanity check. In the 7B reference setting, LBW-Guard reduces final perplexity from 13.21 to 10.74, an 18.7\% improvement, while reducing end-to-end time from 392.54s to 357.02s, a 1.10$\times$ speedup. Under stronger learning-rate stress, AdamW degrades to 1885.24 final perplexity at $LR=3\times10^{-3}$ and 659.76 at $LR=10^{-3}$, whereas LBW-Guard remains trainable at 11.57 and 10.33, respectively. Gradient-clipping baselines do not reproduce this effect.

These results support a scoped systems conclusion that stability-sensitive LLM training can benefit from a governance plane above the optimizer. LBW-Guard provides evidence that bounded runtime control can preserve productive compute under stress while remaining distinct from optimizer replacement and local gradient suppression.
\end{abstract}


\section{Introduction}
Modern model training is increasingly expensive, fragile, and operationally difficult across model scales. Instability is not exclusive to frontier training: smaller and medium-scale training workloads can also exhibit brittle trajectories under aggressive learning rates, longer budgets, wider training modules, or unfavorable batch and sequence regimes. What changes with scale is the operational and economic consequence. As model size, training duration, and infrastructure complexity increase, a failed or severely degraded run consumes more accelerator time, delays experimentation, and increases recovery burden \citep{kaplan2020scaling, hoffmann2022training, chowdhery2023palm}.

The dominant response to training difficulty has historically been optimizer-centric. Adaptive methods such as Adam, refinements such as AdamW, and memory-efficient methods such as Adafactor have made modern deep learning practical \citep{kingma2015adam, loshchilov2019decoupled, shazeer2018adafactor}. However, optimizer-centric abstractions are incomplete when training becomes a fragile runtime process. Training instability has long been studied in feedforward networks, recurrent networks, and adaptive optimization, where gradient-flow pathologies, exploding or vanishing gradients, poor initialization, and non-convergence can impair learning \citep{glorot2010understanding, pascanu2013difficulty, reddi2018convergence}.

Large-model reports make the operational cost of these issues visible. PaLM reported repeated loss spikes and mitigation through checkpoint rollback and skipped batches; OPT reported divergences handled by lowering the learning rate and restarting from earlier checkpoints; GLM-130B reported engineering challenges around loss spikes and divergence \citep{chowdhery2023palm, zhang2022opt, zeng2023glm}. Datacenter studies further show that LLM development is entangled with hardware failure, scheduling complexity, and recovery engineering \citep{hu2024characterization}. Jiang et al.~\citep{jiang2025l4} analyze 428 large-scale LLM training failures from a production platform and report that such failures waste resources and time.

This paper argues that LLM training should be understood as both an optimization process and a runtime control problem. Optimizers compute parameter updates, but they do not by themselves provide a governance layer for detecting, interpreting, and responding to unstable operating conditions during training. This distinction becomes important when training runs are long, costly, and fragile: the central issue is not only whether an optimizer can reduce loss under stable conditions, but whether the training process remains productive when instability emerges.

We instantiate this view through Learn-by-Wire Guard (LBW-Guard), a bounded autonomous training-control governance layer over AdamW. By analogy to fly-by-wire systems in aerospace, learn-by-wire mediates execution in response to runtime conditions while leaving the underlying actuator intact. In this paper, AdamW remains the optimizer; LBW-Guard provides the sensing, regime interpretation, bounded control posture, actuation interface, and telemetry needed to govern optimizer execution under stress.

The contributions of this paper are fivefold: (i) we introduce training-control governance as a systems layer above optimizer execution; (ii) we specify LBW-Guard at the component-control level; (iii) we evaluate it through a Qwen2.5-7B-centered stress-and-robustness suite with 3B and 14B model-size comparisons; (iv) we test whether the observed effect is reducible to ordinary gradient clipping; and (v) we provide preliminary evidence that the effect is not structurally dependent on LoRA through a no-LoRA TinyLlama-1B sanity check.

\section{Related Work}

Large-scale neural-network training has traditionally been organized around the optimizer as the central abstraction for learning. Stochastic optimization methods and their adaptive variants provide the computational mechanism through which model parameters are updated in response to gradient information. Adam, AdamW, Adafactor, AdEMAMix and related methods improve update computation, adaptive scaling, memory efficiency, and regularization behavior \citep{kingma2015adam, loshchilov2019decoupled, shazeer2018adafactor, bottou2018optimization, duchi2011adaptive, pagliardini2025ademamix}. Recent optimizer-benchmarking work further shows that optimizer performance in LLM training must be evaluated under controlled settings that vary model size, batch size, training duration, and optimization regime \citep{semenov2025benchmarking}. This paper builds on that optimizer foundation but does not propose a new update rule. Instead, LBW-Guard introduces a bounded training-control governance layer above AdamW execution. AdamW remains responsible for parameter updates, while LBW-Guard monitors the training state, interprets instability, and applies bounded control over optimizer execution.

A second body of work studies training instability and stabilization. Classical neural-network training research has examined exploding and vanishing gradients, poor signal propagation, initialization sensitivity, and recurrent-network instability \citep{glorot2010understanding, pascanu2013difficulty}. More recent work has analyzed convergence failures and pathological behavior in adaptive optimization, including conditions under which Adam-like methods may fail to converge or behave unstably \citep{reddi2018convergence}. Other stabilization approaches address instability through mechanisms such as gradient clipping, normalization strategies, architectural modifications, or normalization-free training \citep{brock2021high}. These methods are important because they reduce specific sources of instability inside the model or optimizer pipeline. However, they generally intervene locally: they modify gradients, architectures, normalization behavior, or optimizer dynamics. \lbw{} addresses a different level of abstraction. It treats instability as a runtime training condition to be sensed, interpreted, and governed through bounded control over optimizer execution.

Recent LLM-specific studies further motivate the need for training-control perspectives. Large-scale language-model training is known to exhibit loss spikes, divergence, instability under aggressive learning rates, and sensitivity to training configuration. Public reports from PaLM, OPT, and GLM-130B describe practical mitigation strategies such as checkpoint rollback, skipped batches, learning-rate reduction, and restart from earlier checkpoints \citep{chowdhery2023palm, zhang2022opt, zeng2023glm}. These examples show that instability is not merely a theoretical optimization concern; it becomes an operational event that must be detected, interpreted, and managed during training. Recent work on Adam instability and loss-spike mitigation in large language models also reinforces the view that instability can emerge dynamically during training and may require mechanisms beyond ordinary optimizer selection \citep{molybog2023theory, takase2025spike}. \lbw{} is positioned in this gap: it does not replace optimizer research, but adds a run-level governance plane that can respond to instability while preserving the underlying optimizer.

Infrastructure and production studies make this problem economically significant. As models become larger and training runs become longer, instability affects not only final loss but also compute productivity, wall-clock time, engineering effort, and experiment reliability. Scaling-law and compute-optimal training studies have shown that model quality is closely tied to compute allocation, data scale, and training efficiency \citep{kaplan2020scaling, hoffmann2022training}. In production environments, however, compute is not consumed only by successful learning; it is also consumed by failed runs, degraded trajectories, hardware failures, scheduling inefficiencies, checkpoint recovery, and operational restarts. Datacenter studies show that large-language-model development is entangled with infrastructure failures, resource imbalance, and fault-tolerant recovery \citep{hu2024characterization}. Production evidence from large-scale training platforms further shows that training failures can waste substantial resources and time \citep{jiang2025l4}. These findings motivate evaluation criteria that go beyond final validation loss alone. A training method should also be assessed by whether it preserves productive compute under stress.

This distinction is central to the present paper. Standard optimizer comparisons typically ask which update rule achieves better loss or convergence under a given configuration. Stabilization methods often ask whether a specific pathology can be reduced through clipping, normalization, or architectural adjustment. \lbw{} asks a complementary systems question: can a training process be governed during execution so that instability is sensed early, interpreted as an operating condition, and handled through bounded corrective action? This shifts the focus from optimizer replacement to training-control governance. The optimizer still computes updates; the governance layer manages the conditions under which those updates are executed.

The closest conceptual analogy is therefore not another optimizer, but a control layer around an existing execution mechanism. In the same way that safety-critical engineered systems often separate the actuator from the control logic that governs its operation, \lbw{} separates \adamw{} as the optimizer actuator from the bounded governance logic that monitors and modulates training execution. This architectural separation is important because it allows the method to remain compatible with existing optimizer infrastructure while adding telemetry, regime interpretation, bounded control posture, and logging. The logger also makes the control process observable through quantities such as control-active steps, regime switches, scale, and control energy. This observability supports a systems-level interpretation of training behavior rather than treating the method as an opaque performance improvement.

The empirical comparison with gradient clipping is especially important for positioning. Gradient clipping is a widely used stabilization technique, but it acts primarily as local gradient suppression. It does not by itself constitute a training-state governance loop: it does not interpret operating regimes, distinguish stress from recovery-like conditions, or record control-active behavior as a run-level process. The clipping baselines in this paper therefore test whether \lbw{}'s effect can be reduced to ordinary gradient magnitude limitation. The results suggest that clipping alone does not reproduce the observed trainability preservation under stress, supporting the claim that \lbw{} operates at a different level of abstraction.

In summary, prior work provides three foundations for this paper: optimizer research explains how parameter updates are computed; stabilization research explains how local training pathologies can be reduced; and production infrastructure studies explain why instability has operational and economic consequences. \lbw{} contributes to the space between these literatures. It treats LLM training as a runtime system that requires not only optimization, but also bounded autonomous control governance over the training process. This framing motivates the component-level method specification and the stress-and-robustness evaluation that follow.

\begin{figure}[t]
\centering
\resizebox{0.95\linewidth}{!}{%
\begin{tikzpicture}[
  node distance=1.6cm,
  box/.style={draw, rounded corners, align=center, minimum width=2.9cm, minimum height=1.0cm, font=\small},
  smallbox/.style={draw, rounded corners, align=center, minimum width=3.1cm, minimum height=0.9cm, font=\small},
  arrow/.style={-Latex, thick}
]
\node[box] (sensor) {Sensor\\\footnotesize lightweight telemetry};
\node[box, right=of sensor] (analyzer) {Analyzer\\\footnotesize run-state estimate};
\node[box, right=of analyzer] (policy) {Policy / Controller\\\footnotesize bounded posture};
\node[box, right=of policy] (actuator) {Actuator\\\footnotesize governs AdamW execution};
\node[smallbox, below=1.6cm of analyzer] (adamw) {AdamW optimizer\\\footnotesize update engine};
\node[smallbox, below=1.6cm of policy] (logger) {Telemetry logger\\\footnotesize control-active steps};

\draw[arrow] (sensor) -- (analyzer);
\draw[arrow] (analyzer) -- (policy);
\draw[arrow] (policy) -- (actuator);
\draw[arrow] (actuator.south) |- (adamw.east);
\draw[arrow] (adamw.west) -| (sensor.south);
\draw[arrow] (policy.south) -- (logger.north);
\draw[arrow] (actuator.south) -- (logger.east);
\draw[arrow] (logger.west) -- (adamw.east);
\end{tikzpicture}}
\caption{\lbw{} architecture. \adamw{} remains the optimizer plane, while \lbw{} operates as a bounded control-governance plane over training execution.}
\label{fig:architecture}
\end{figure}
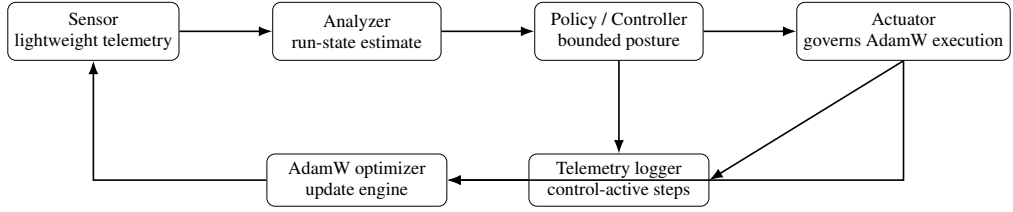

\section{\lbw: Component-Control Method}
\lbw{} is a bounded autonomous control-governance layer that wraps \adamw{} without redefining \adamw. The optimizer plane remains responsible for parameter updates. The governance plane monitors the training trajectory, interprets the operating condition, selects a bounded control posture, and applies constrained actuation to the optimizer execution path. Figure~\ref{fig:architecture} summarizes the architecture.

\begin{table}[t]
\centering
\small
\begin{tabular}{p{0.18\linewidth}p{0.25\linewidth}p{0.45\linewidth}}
\toprule
\textbf{Component} & \textbf{Role} & \textbf{Public specification} \\
\midrule
Sensor & Collects telemetry & Loss trajectory, ratio/trend signals, optional lightweight probes \\
Analyzer & Interprets training condition & Stable, stress, spike/oscillation, or recovery-like regimes \\
Policy/controller & Selects bounded posture & Constrained scale/damping/release under predefined limits \\
Actuator & Applies bounded control & Modulates AdamW execution without replacing the update rule \\
Logger & Records control behavior & Control-active steps, regime switches, scale, control energy \\
\bottomrule
\end{tabular}
\caption{Component-level method specification for \lbw.}
\label{tab:components}
\end{table}

The sensing layer is read-only and is not structurally tied to LoRA. The sensor can use lightweight loss-only telemetry or sparse probing; full-gradient instrumentation is optional rather than required. The contribution is a systems architecture for bounded run-level governance over existing optimizers; the present paper evaluates one implementation through observable telemetry and empirical behavior.

Table~\ref{tab:components} specifies \lbw{} at the component-control level. The table is intended to clarify the public methodological boundary of the system: \lbw{} is not presented as a new optimizer update rule, but as a bounded training-control governance layer that operates above \adamw. \adamw{} remains responsible for computing parameter updates, while \lbw{} observes the training process, interprets the current operating condition, and applies constrained control to the execution path.

The component structure follows a sensing--interpretation--policy--actuation--logging loop. The sensor collects lightweight training telemetry, such as loss trajectory, ratio or trend signals, and optional probes. The analyzer converts these signals into interpretable training conditions, including stable, stressed, spike/oscillation, or recovery-like regimes. The policy/controller then selects a bounded control posture under predefined limits, ensuring that the system can dampen or release control without changing the fixed training objective. The actuator applies this bounded posture to \adamw{} execution, modulating how the optimizer is executed rather than replacing the \adamw{} update rule itself. Finally, the logger records observable control behavior, including control-active steps, regime switches, scale values, and control energy.

This component-level specification serves two purposes. First, it makes the method empirically interpretable: the reported results can be connected to observable control behavior rather than treated as a black-box training improvement. Second, it preserves the distinction between reproducible scientific specification and proprietary controller implementation. The paper therefore discloses the architectural roles, public telemetry categories, bounded-control interface, and logged evidence needed to evaluate the claim, while avoiding unnecessary disclosure of implementation-specific policy logic.

\paragraph{Component-control loop.}
At each training step, the sensor collects lightweight training-state telemetry; the analyzer updates recent state and assigns an operating condition; the policy/controller selects a bounded control posture under predefined limits; the actuator applies bounded scale/damping/release to the \adamw{} execution path; and the logger records control-active steps, regime switches, stress mode, scale, and control energy.

\section{Experimental Design}
We evaluate LBW-Guard in controlled single-GPU LLM training settings using Qwen2.5 model variants, WikiText-103 raw, CUDA, PyTorch AdamW, and LoRA-based training stress tests. Detailed base-run settings, including model variants, dataset split, training steps, sequence length, LoRA configuration, and clipping baselines, are reported in Appendix~\ref{app:settings}. The purpose of the experimental design is not to claim frontier-scale pretraining validation, but to test whether bounded training-control governance changes training behavior under instability-sensitive conditions. The experiments therefore emphasize stress, robustness, and comparative behavior against AdamW rather than absolute state-of-the-art language-model performance.

\begin{table}[t]
\centering
\small
\begin{tabular}{p{0.25\linewidth}p{0.29\linewidth}p{0.38\linewidth}}
\toprule
\textbf{Axis} & \textbf{Settings} & \textbf{Purpose} \\
\midrule
Model size & 3B, 7B, 14B & Tests whether the effect survives scale changes \\
Learning rate & $3\times 10^{-3}$, $10^{-3}$, $5\times 10^{-4}$ & Tests stress under aggressive and moderate LR \\
Gradient clipping & $g=1.0$, $g=0.5$ & Tests local gradient suppression \\
No-LoRA sanity check & TinyLlama 1B full-parameter run & Tests whether the effect is tied to adapters \\
Long-budget stress & 5000-step run & Tests productive-compute preservation \\
Seed repeatability & Seeds 7, 42, 123 & Checks random-path sensitivity \\
\bottomrule
\end{tabular}
\caption{Stress-and-robustness suite used in the paper.}
\label{tab:suite}
\end{table}

The central comparison is between standard AdamW and AdamW executed under LBW-Guard. In all matched comparisons, AdamW remains the underlying optimizer. LBW-Guard does not replace the AdamW update rule; it wraps the training process with a bounded control-governance layer that senses training state, interprets instability, and modulates optimizer execution under predefined limits. This design isolates the contribution of the governance layer from the contribution of the optimizer itself. The empirical question is therefore whether adding bounded control over optimizer execution improves trainability, final perplexity, and productive compute under stress.

The public optimizer-construction interface is reported in Appendix~\ref{app:execution_interface} to clarify that LBW-Guard preserves the AdamW-facing hyperparameters while adding bounded control-governance arguments.

The primary rerun uses Qwen2.5-7B as the empirical anchor. This setting is chosen because it is large enough to expose nontrivial training instability and runtime cost, while remaining feasible for controlled repeated experiments. We then include Qwen2.5-3B and Qwen2.5-14B model-size comparisons to test whether the observed effect is specific to one model scale or persists across smaller and larger presets. These model-size experiments are not intended as a scaling-law study; instead, they serve as a robustness check for the component-level claim that LBW-Guard can improve stability-sensitive training behavior across different model sizes.

The dataset is WikiText-103 raw, using the full training split for training and the full validation split for evaluation. Final validation perplexity is the main quality metric because it provides a direct measure of language-model fit under the same data and evaluation protocol. We also report final loss, wall-clock time, tokens per second, end-to-end speedup, and LBW-Guard telemetry where applicable. Runtime metrics are included because the paper studies training control governance as a systems problem: a method that preserves trainability under stress should also be evaluated by how much compute remains productive.

The learning-rate stress suite is a core part of the design. We evaluate aggressive and moderate learning-rate conditions, including $3 \times 10^{-3}$, $1 \times 10^{-3}$, and $5 \times 10^{-4}$. These settings intentionally include regimes where AdamW may become unstable or severely degraded. The purpose is not to recommend these learning rates as default recipes, but to evaluate whether LBW-Guard can preserve trainability when the optimization trajectory becomes fragile. This stress-test framing is important because bounded governance is most relevant when training is close to failure, not only when training is already stable.

We also include gradient-clipping baselines at $LR=10^{-3}$ to test whether LBW-Guard's effect can be explained by ordinary local gradient suppression. Gradient clipping is a common stabilization technique, but it does not constitute a run-level control-governance loop. It limits gradient magnitude, whereas LBW-Guard senses the training trajectory, assigns operating regimes, selects bounded control postures, and records control-active behavior. The clipping comparison therefore asks whether a simpler stabilization mechanism can reproduce the observed trainability and perplexity improvements.

To test whether the effect is tied only to LoRA-based adapter training, we include a no-LoRA TinyLlama 1B full-parameter sanity check. This experiment is intentionally scoped as a sanity check rather than a full pretraining benchmark. Its role is to examine whether LBW-Guard's bounded control-governance behavior can remain useful when the training process is not structurally dependent on LoRA adapters. A positive result in this setting strengthens the interpretation that LBW-Guard is a training-execution governance method rather than merely a LoRA-specific adjustment.

The experimental suite also records telemetry from the LBW-Guard control process. These telemetry signals include control-active steps, regime switches, scale values, and control energy. Such measurements are important because they make the governance layer observable. Without telemetry, an improvement in perplexity could be treated as a black-box effect. With telemetry, the paper can connect empirical outcomes to control behavior during training, supporting the interpretation that LBW-Guard changes the trajectory of the run under stress.

Table~\ref{tab:suite} summarizes the stress-and-robustness suite used in the paper. The design combines model-size variation, learning-rate stress, clipping baselines, a no-LoRA sanity check, long-budget stress, and seed repeatability evidence. Together, these experiments test whether LBW-Guard improves training behavior across several instability-sensitive conditions while keeping the claim appropriately scoped to controlled LLM training stress tests.

\section{Results}

This section reports the empirical behavior of LBW-Guard across the stress-and-robustness suite. The results are organized around six questions: whether LBW-Guard improves the main Qwen2.5-7B reference setting; whether the effect persists across model-size presets; whether it preserves trainability under learning-rate stress; whether the gains can be explained by ordinary gradient clipping; whether the effect is structurally tied to LoRA; and whether preliminary seed evidence suggests repeatability. Across these experiments, the main evaluation metric is final validation perplexity, supplemented by final loss, wall-clock time, end-to-end speedup, and LBW-Guard control telemetry where available.

\subsection{Qwen2.5-7B Reference Setting}

The empirical anchor is the Qwen2.5-7B setting. This setting is used as the main reference point because it is large enough to expose meaningful training instability and runtime cost, while remaining feasible for controlled single-GPU stress testing. As shown in table~\ref{tab:modelsize}, LBW-Guard reduces final evaluation perplexity by 18.7\%, from 13.2086 under AdamW to 10.7353 under LBW-Guard. This improvement indicates that the governance layer does not merely preserve training from collapse; it also improves the quality of the final training trajectory under the matched configuration.

The runtime behavior is also favorable. LBW-Guard reduces end-to-end time from 392.54 seconds to 357.02 seconds, corresponding to a 1.10$\times$ speedup. This result is important because LBW-Guard adds a control layer above AdamW, which might be expected to introduce overhead. The observed improvement therefore suggests a trajectory-efficiency effect: by dampening unstable or inefficient training behavior, LBW-Guard can reduce wasted training motion and preserve more productive compute. Since AdamW remains the optimizer and LBW-Guard does not replace the update rule, this result supports the paper's central claim that optimizer execution can be improved through bounded run-level governance.

The telemetry further supports this interpretation at tables~\ref{tab:reference7b}. In the Qwen2.5-7B reference setting, the controller records 991 control-active steps and 29 regime switches. This indicates that LBW-Guard is not a passive wrapper around AdamW. It actively monitors the training process, interprets changing operating conditions, and adjusts its bounded control posture throughout the run. The presence of regime switches is particularly relevant because it suggests that the training trajectory moves through multiple operating states rather than remaining in a single static condition. The results therefore connect the quality and runtime improvements to observable control behavior rather than treating the outcome as an unexplained black-box gain.

\begin{table}[t]
\centering
\small
\begin{tabular}{lrrrrr}
\toprule
\textbf{Method} & \textbf{Loss} & \textbf{PPL} & \textbf{Tokens/s} & \textbf{Wall (s)} & \textbf{E2E} \\
\midrule
AdamW & 2.5809 & 13.2086 & 621.98 & 392.54 & 1.000$\times$ \\
LBW-Guard & 2.3735 & 10.7353 & 734.84 & 357.02 & 1.099$\times$ \\
\bottomrule
\end{tabular}
\caption{Qwen2.5-7B size-preset reference setting. \lbw{} reports 991 control-active steps.}
\label{tab:reference7b}
\end{table}

\subsection{Model-Size Robustness}

\begin{figure}[t]
\centering
\begin{tikzpicture}
\begin{axis}[
  ybar,
  width=0.84\linewidth,
  height=0.62\linewidth,
  ylabel={Final eval perplexity},
  xlabel={Model size},
  symbolic x coords={3B,7B,14B},
  xtick=data,
  ymin=0,
  legend style={at={(0.98,0.98)},anchor=north east},
  bar width=14pt,
  enlarge x limits=0.25,
  nodes near coords,
  nodes near coords align={vertical},
]
\addplot coordinates {(3B,10.3017) (7B,13.2086) (14B,11.0618)};
\addplot coordinates {(3B,9.6548) (7B,10.7353) (14B,9.0654)};
\legend{AdamW,LBW-Guard}
\end{axis}
\end{tikzpicture}
\caption{Model-size robustness. \lbw{} improves final perplexity across 3B, 7B, and 14B presets; runtime effects vary by scale.}
\label{fig:modelsize}
\end{figure}
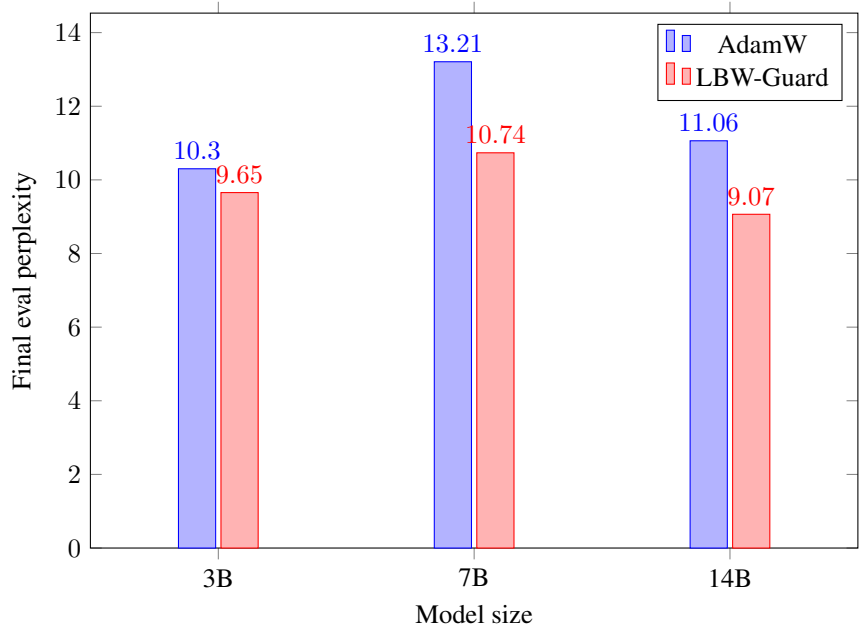

The model-size comparison evaluates whether LBW-Guard's effect is specific to one model preset or whether it persists across different Qwen2.5 scales. As shown in table~\ref{tab:modelsize} and figure~\ref{fig:modelsize}, LBW-Guard improves final evaluation perplexity across the 3B, 7B, and 14B presets. In the 3B setting, final perplexity decreases by 6.3\%, from 10.3017 to 9.6548. In the 7B setting, final perplexity decreases by 18.7\%, from 13.2086 to 10.7353. In the 14B setting, final perplexity decreases by 18.0\%, from 11.0618 to 9.0654.

These results suggest that the effect is not isolated to a single model size. The largest relative gains appear in the 7B and 14B settings, where LBW-Guard improves both final perplexity and runtime. The 3B setting is more quality-led: LBW-Guard improves final perplexity but introduces a small runtime tradeoff, with an end-to-end speedup of 0.967$\times$. This distinction is useful because it prevents overclaiming. LBW-Guard does not universally make every run faster in every configuration; rather, its most consistent effect in the model-size suite is improved final perplexity, while runtime benefits depend on the stress and scale regime.

The model-size results also support the interpretation that LBW-Guard is a training-control method rather than a configuration-specific trick. If the improvement appeared only in one preset, it could be attributed to an accidental interaction between model size, LoRA configuration, or learning-rate schedule. Instead, the consistent perplexity reduction across 3B, 7B, and 14B suggests that bounded control over optimizer execution can improve training behavior across multiple scale conditions. The result remains scoped: this is not a full scaling-law claim, but it is a meaningful robustness check for the component-level method.

\subsection{Learning-Rate Stress}

\begin{figure}[t]
\centering
\begin{tikzpicture}
\begin{semilogyaxis}[
  ybar,
  width=0.84\linewidth,
  height=0.42\linewidth,
  ylabel={Final eval perplexity (log scale)},
  xlabel={Learning rate},
  symbolic x coords={$3\times 10^{-3}$,$10^{-3}$,$5\times 10^{-4}$},
  xtick=data,
  legend style={at={(0.98,0.98)},anchor=north east},
  bar width=14pt,
  enlarge x limits=0.25,
  ymin=8,
]
\addplot coordinates {($3\times 10^{-3}$,1885.24) ($10^{-3}$,659.76) ($5\times 10^{-4}$,11.6625)};
\addplot coordinates {($3\times 10^{-3}$,11.5704) ($10^{-3}$,10.3280) ($5\times 10^{-4}$,10.2582)};
\legend{AdamW,LBW-Guard}
\end{semilogyaxis}
\end{tikzpicture}
\caption{Learning-rate stress curve. Perplexity is shown on a log scale.}
\label{fig:lrstress}
\end{figure}
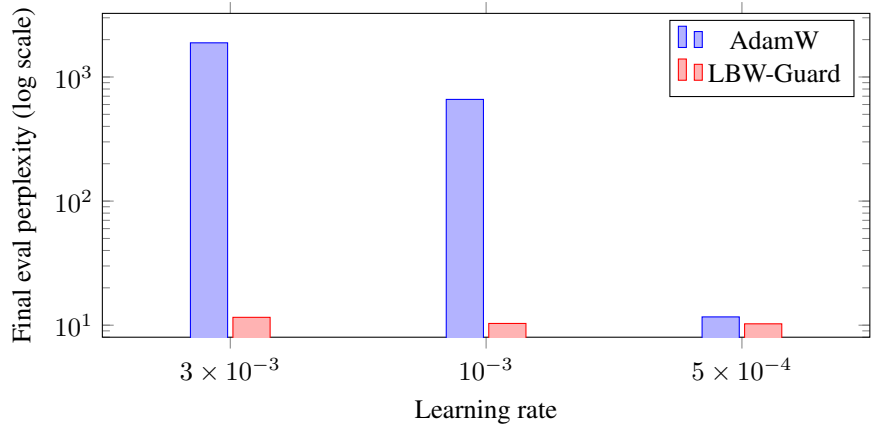

Table~\ref{tab:lrstress} and figure~\ref{fig:lrstress} provides the clearest evidence for LBW-Guard's value under learning-rate stress. At $LR=3 \times 10^{-3}$, AdamW becomes severely degraded, reaching 1885.24 final perplexity, while LBW-Guard remains trainable at 11.5704 and is 1.084$\times$ faster end-to-end. At $LR=10^{-3}$, AdamW also becomes severely degraded and reaches 659.76 final perplexity, while LBW-Guard remains at 10.3280. At $LR=5 \times 10^{-4}$, where the setting is less severely stressed, LBW-Guard still improves final perplexity from 11.6625 to 10.2582 and improves end-to-end speed by 1.016$\times$.

\begin{table}[t]
\centering
\small
\begin{tabular}{lrrrr}
\toprule
\textbf{Setting} & \textbf{AdamW PPL} & \textbf{LBW-Guard PPL} & \textbf{PPL reduction} & \textbf{E2E speedup} \\
\midrule
3B & 10.3017 & 9.6548 & 6.3\% & 0.967$\times$ \\
7B & 13.2086 & 10.7353 & 18.7\% & 1.099$\times$ \\
14B & 11.0618 & 9.0654 & 18.0\% & 1.181$\times$ \\
\bottomrule
\end{tabular}
\caption{Model-size ablation.}
\label{tab:modelsize}
\end{table}

The high-perplexity AdamW results at $LR=3 \times 10^{-3}$ and $LR=10^{-3}$ should be interpreted as failure-sensitive stress cases rather than ordinary recommended training recipes. Their value is diagnostic: they test what happens when the training process is pushed into an unstable or severely degraded operating regime. In these conditions, AdamW continues to consume compute but produces an unusable or heavily degraded final trajectory. LBW-Guard, by contrast, preserves trainability and keeps final perplexity within a usable range. This is the most direct empirical support for the paper's central systems argument: training methods should be evaluated not only by how they behave in stable settings, but also by whether they preserve productive compute when instability emerges.

\begin{table}[t]
\centering
\small
\begin{tabular}{lrrrr}
\toprule
\textbf{LR} & \textbf{AdamW PPL} & \textbf{LBW-Guard PPL} & \textbf{PPL reduction} & \textbf{E2E speedup} \\
\midrule
$3\times 10^{-3}$ & 1885.24 & 11.5704 & 99.4\% & 1.084$\times$ \\
$10^{-3}$ & 659.76 & 10.3280 & 98.4\% & 0.994$\times$ \\
$5\times 10^{-4}$ & 11.6625 & 10.2582 & 12.0\% & 1.016$\times$ \\
\bottomrule
\end{tabular}
\caption{Learning-rate stress.}
\label{tab:lrstress}
\end{table}

The moderate learning-rate condition at $LR=5 \times 10^{-4}$ is also important. In this setting, AdamW does not collapse, yet LBW-Guard still improves final perplexity by 12.0\%. This suggests that LBW-Guard's contribution is not limited to rescuing failed runs. It can also improve the efficiency of a trainable trajectory by governing local instability, stress, or inefficient motion that may not appear as outright divergence. The distinction between ``preventing failure'' and ``improving a trainable trajectory'' strengthens the empirical case for training control governance.

A natural alternative explanation is that LBW-Guard simply behaves like a lower fixed learning rate. The available learning-rate sweep weakens the simplest version of this objection. AdamW at $LR=5 \times 10^{-4}$ reaches final perplexity 11.6625, whereas LBW-Guard at $LR=10^{-3}$ reaches 10.3280 and LBW-Guard at $LR=5 \times 10^{-4}$ reaches 10.2582. If LBW-Guard were merely equivalent to lowering the learning rate, then AdamW at the lower learning rate would be expected to close most of the gap. It does not. This does not fully exhaust all possible tuned learning-rate explanations, but it shows that the observed effect is not reducible to the simplest fixed-LR interpretation.

\begin{table}[t]
\centering
\small
\begin{tabular}{lrrrl}
\toprule
\textbf{Method} & \textbf{Final PPL} & \textbf{Final loss} & \textbf{Wall (s)} & \textbf{Interpretation} \\
\midrule
AdamW + clip $g=1.0$ & 659.76 & 6.4919 & 513.07 & severe degradation \\
AdamW + clip $g=0.5$ & 891.37 & 6.7928 & 527.47 & severe degradation \\
LBW-Guard + clip $g=1.0$ & 10.3936 & 2.3412 & 476.35 & trainable \\
\bottomrule
\end{tabular}
\caption{Gradient-clipping baseline at LR=$10^{-3}$.}
\label{tab:clipping}
\end{table}

\begin{figure}[t]
\centering
\begin{tikzpicture}
\begin{semilogyaxis}[
  ybar,
  width=0.84\linewidth,
  height=0.42\linewidth,
  ylabel={Final eval perplexity (log scale)},
  symbolic x coords={AdamW clip 1.0,AdamW clip 0.5,LBW-Guard clip 1.0},
  xtick=data,
  x tick label style={rotate=15,anchor=east},
  bar width=24pt,
  ymin=8,
  nodes near coords,
]
\addplot coordinates {(AdamW clip 1.0,659.76) (AdamW clip 0.5,891.37) (LBW-Guard clip 1.0,10.3936)};
\end{semilogyaxis}
\end{tikzpicture}
\caption{Gradient-clipping baseline at LR=$10^{-3}$. Clipping alone does not reproduce \lbw{}'s gains.}
\label{fig:clipping}
\end{figure}
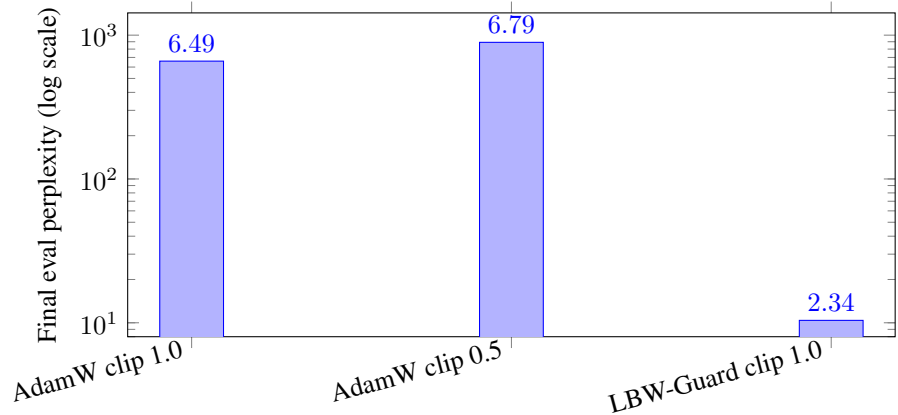

\subsection{Gradient-Clipping Baseline}

We next test whether ordinary gradient clipping explains LBW-Guard's gains. This comparison is important because gradient clipping is a common and simple stabilization technique. If LBW-Guard's benefits were merely the result of suppressing large gradients, then clipped AdamW should produce similar improvements under the same stress condition.

The results shown in table~\ref{tab:clipping} and figure~\ref{fig:clipping} do not support that explanation. In the $LR=10^{-3}$ clipping suite, AdamW with global clipping at $g=1.0$ reaches final perplexity 659.76, while AdamW with stricter clipping at $g=0.5$ reaches 891.37. Both results remain severely degraded. By contrast, LBW-Guard with $g=1.0$ remains trainable at 10.3936 and is 1.08$\times$ faster end-to-end. The stricter AdamW clipping condition does not improve stability; it worsens the final result, suggesting that naive gradient suppression can damage useful learning dynamics without solving the underlying trajectory problem.

This comparison supports the claim that LBW-Guard operates at a different level of abstraction from gradient clipping. Clipping limits gradient magnitude locally, but it does not sense training regimes, distinguish stress from recovery-like conditions, select a bounded control posture, or log control-active behavior over the run. LBW-Guard instead treats instability as a runtime operating condition. The empirical gap between clipped AdamW and LBW-Guard therefore strengthens the interpretation that the improvement comes from bounded control governance rather than from simple gradient magnitude limitation.

\begin{table}[t]
\centering
\small
\begin{tabular}{lrrrl}
\toprule
\textbf{Method} & \textbf{Final PPL} & \textbf{Final loss} & \textbf{Wall (s)} & \textbf{Interpretation} \\
\midrule
AdamW + clip $g=1.0$ & 319.67 & 5.7673 & 276.75 & severe degradation \\
LBW-Guard + clip $g=1.0$ & 18.5470 & 2.9203 & 245.68 & trainable \\
AdamW + clip $g=0.5$ & 428.04 & 6.0592 & 281.92 & worse degradation \\
\bottomrule
\end{tabular}
\caption{No-LoRA TinyLlama 1B full-parameter sanity check.}
\label{tab:nolora}
\end{table}

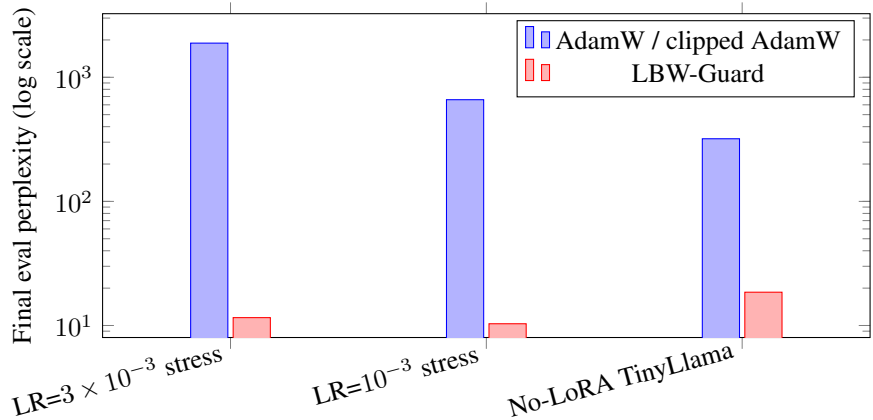
\begin{figure}[t]
\centering
\begin{tikzpicture}
\begin{semilogyaxis}[
  ybar,
  width=0.84\linewidth,
  height=0.42\linewidth,
  ylabel={Final eval perplexity (log scale)},
  symbolic x coords={LR=$3\times 10^{-3}$ stress,LR=$10^{-3}$ stress,No-LoRA TinyLlama},
  xtick=data,
  x tick label style={rotate=15,anchor=east},
  legend style={at={(0.98,0.98)},anchor=north east},
  bar width=14pt,
  enlarge x limits=0.25,
  ymin=8,
]
\addplot coordinates {(LR=$3\times 10^{-3}$ stress,1885.24) (LR=$10^{-3}$ stress,659.76) (No-LoRA TinyLlama,319.67)};
\addplot coordinates {(LR=$3\times 10^{-3}$ stress,11.5704) (LR=$10^{-3}$ stress,10.3280) (No-LoRA TinyLlama,18.5470)};
\legend{AdamW / clipped AdamW,LBW-Guard}
\end{semilogyaxis}
\end{tikzpicture}
\caption{Failure-sensitive scenarios. \adamw{} consumes compute but reaches unusable final perplexity in stress cases, while \lbw{} preserves trainability.}
\label{fig:failuresensitive}
\end{figure}

\subsection{No-LoRA Full-Parameter Sanity Check}

To test whether the effect is specific to adapter-based training, we additionally evaluate a no-LoRA TinyLlama 1B full-parameter setting. This experiment is intentionally scoped as a sanity check rather than a full frontier-scale pretraining benchmark. Its role is to examine whether the bounded control-governance effect is structurally dependent on LoRA adapters.

The result shown in table~\ref{tab:nolora} and figure~\ref{fig:failuresensitive} suggests that it is not. Under the same $g=1.0$ clipping condition, AdamW reaches final evaluation perplexity 319.67, whereas LBW-Guard reaches 18.5470. LBW-Guard also reduces wall-clock time from 276.75 seconds to 245.68 seconds. AdamW with stricter clipping at $g=0.5$ degrades further to 428.04 final perplexity. This pattern is consistent with the LoRA-based stress tests: naive clipping does not preserve trainability, while LBW-Guard maintains a substantially more usable trajectory.

The no-LoRA result should be interpreted carefully. It does not prove that LBW-Guard will generalize to all full-parameter pretraining settings, nor does it substitute for large-scale multi-GPU pretraining validation. However, it does provide useful evidence that the observed behavior is not purely an artifact of LoRA adapters. This matters for the method claim because LBW-Guard is framed as a run-level control-governance layer over optimizer execution, not as a LoRA-specific stabilization mechanism.

\subsection{Seed Repeatability}

The available seed evidence remains limited but useful. In the prior 3B seed comparison, AdamW obtains mean final perplexity $12.68 \pm 0.14$, whereas LBW-Guard obtains $9.69 \pm 0.06$ across seeds 7, 42, and 123. The lower mean indicates a consistent quality advantage for LBW-Guard in this setting, while the smaller standard deviation suggests that the governed training trajectory may also be less sensitive to random-path variation.

We treat this result as preliminary repeatability evidence rather than complete statistical validation. The seed comparison is useful because it reduces the likelihood that the observed improvement is entirely due to a single favorable random run. However, broader confidence intervals, additional seeds, multiple datasets, and larger-scale training environments are still needed before making strong statistical generalization claims. For the present paper, the seed evidence supports the more modest conclusion that LBW-Guard's advantage appears repeatable in the tested 3B setting.

\subsection{Summary of Empirical Findings}

Taken together, the results support four main findings. First, in the Qwen2.5-7B reference setting, LBW-Guard improves both final perplexity and end-to-end runtime while producing observable control telemetry. Second, across Qwen2.5 model-size presets, LBW-Guard consistently improves final perplexity, with the strongest gains in the 7B and 14B settings. Third, under learning-rate stress, LBW-Guard preserves trainability in regimes where AdamW becomes severely degraded. Fourth, gradient clipping does not reproduce the effect, and the no-LoRA sanity check suggests that the method is not structurally dependent on adapter-based training.

These findings are consistent with the paper's central claim: LBW-Guard should be understood as a bounded autonomous training control-governance layer rather than an optimizer replacement. AdamW computes the parameter update, while LBW-Guard governs the runtime execution conditions under which AdamW operates. The empirical contribution is therefore not only lower perplexity in selected runs, but evidence that bounded control governance can preserve productive training behavior under instability-sensitive conditions.

\paragraph{Reproducibility artifact.}
The reproducibility artifact is intended to support methodological transparency and partial reproducibility, consistent with broader recommendations for improving reproducibility in machine learning research \citep{pineau2021improving}. To support inspection and partial reproducibility, we provide a lightweight Colab-based reproduction artifact archived on Zenodo \citep{radianis2026zenodo} . This artifact is not the full internal experiment code used to generate every reported result. Instead, it is a simplified execution workflow intended to help readers reproduce the structure of the AdamW versus LBW-Guard comparison, inspect the configuration pathway, and validate the main evaluation procedure.

Because the artifact runs in a public Colab environment, results may differ from the reported paper results due to hardware variation, runtime availability, library versions, random seeds, CUDA behavior, memory constraints, and implementation differences between the lightweight artifact and the original controlled experiment environment. The LBW-Guard controller implementation is proprietary, the artifact should be interpreted as partial reproducibility support rather than a full open-source release of the internal control policy. The paper therefore discloses the component-level method specification, public run settings, result tables, telemetry categories, and external execution reference needed to evaluate the empirical claims.

\section{Discussion and Limitations}

From the experiment, the main empirical pattern is that LBW-Guard changes the behavior of training under stress. It is not a new optimizer and does not introduce a new parameter-update rule. AdamW still computes the update, while LBW-Guard governs the runtime conditions under which that update is executed. This distinction is central to the interpretation of the results. The model-size experiments show consistent quality gains across the 3B, 7B, and 14B Qwen2.5 presets. The learning-rate stress suite shows that LBW-Guard preserves trainability in regimes where AdamW becomes severely degraded. The clipping baselines show that ordinary global gradient clipping does not reproduce the effect. Taken together, these results support the view that training-control governance is not simply another optimizer variant or another local stabilization trick, but a distinct systems layer above optimization.

The strongest evidence comes from the learning-rate stress setting. Under aggressive learning rates, AdamW continues to consume compute but produces unusable or severely degraded final perplexity. LBW-Guard, by contrast, remains within a trainable range. This pattern matters because modern LLM training is not only an optimization problem; it is also an operational process in which instability can waste accelerator time, delay experimentation, and require manual intervention. A method that preserves trainability under stress therefore has value beyond final perplexity alone. It improves the probability that a run remains productive when the training trajectory becomes fragile.

The model-size results suggest that the effect is not isolated to a single model preset. LBW-Guard improves final perplexity across 3B, 7B, and 14B settings, with larger relative gains in the 7B and 14B cases. This does not establish a scaling law, and it should not be interpreted as proof that the method will behave identically at frontier scale. However, it does indicate that the observed behavior is not merely an accidental outcome of one model size. The result is better understood as robustness evidence for the component-level claim: bounded control over optimizer execution can improve stability-sensitive training behavior across multiple model-scale conditions.

The end-to-end speed improvements in the 7B and 14B settings may appear counterintuitive because LBW-Guard adds a control layer above AdamW. We interpret these gains as a trajectory-efficiency effect rather than a raw per-step overhead effect. A control layer can introduce some computational overhead, but it can also reduce inefficient training motion, severe instability, and recovery burden. In other words, the relevant question is not only whether each step is marginally cheaper, but whether the overall trajectory uses compute productively. Because the public controller configuration uses $c_{\max}=1.0$, LBW-Guard cannot obtain speedups by amplifying optimizer steps beyond the base execution path. The observed runtime gains are therefore better interpreted as improved training trajectory efficiency rather than more aggressive update scaling.

The clipping results are especially important for interpreting mechanism. Gradient clipping is a reasonable baseline because it is a common response to instability. However, clipping acts locally by limiting gradient magnitude. It does not observe the training trajectory as a runtime system, classify operating regimes, adjust control posture dynamically, or log control-active behavior. In the reported stress setting, clipped AdamW remains severely degraded, while LBW-Guard remains trainable. This does not imply that gradient clipping is useless; clipping remains valuable in many training recipes. Rather, the result shows that LBW-Guard's behavior cannot be reduced to naive gradient suppression. The method appears to operate at a higher level of abstraction: it governs optimizer execution based on the state of the training process.

The no-LoRA TinyLlama sanity check also strengthens the interpretation, although it remains limited. Because most experiments use LoRA-based training, one concern is that the observed effect may be adapter-specific. The no-LoRA full-parameter result weakens that concern by showing that LBW-Guard can preserve trainability outside the LoRA setting. However, this experiment is still only a sanity check. It does not replace large-scale full-parameter pretraining experiments, multi-GPU distributed training validation, or broader evaluation across architectures and datasets. Its role is narrower: it suggests that the control-governance effect is not structurally dependent on LoRA adapters.

The method also reframes what should be observed during training. A standard optimizer experiment often reports final loss, final perplexity, throughput, and perhaps memory footprint. A training-control governance experiment should additionally report whether control was active, how frequently operating regimes changed, whether control remained within predefined limits, and how much control effort was applied. This telemetry does not replace final quality metrics, but it improves interpretability by connecting model outcomes to runtime control behavior. In this paper, control-active steps and regime switches provide evidence that LBW-Guard is not merely a passive wrapper. It actively monitors and modulates the training process throughout the run.

This observability is important for scientific and engineering reasons. Scientifically, telemetry helps distinguish a genuine control-governance effect from an unexplained performance improvement. Engineering-wise, telemetry gives practitioners a way to inspect whether the training process is stable, stressed, oscillatory, or recovering. In future training systems, such information may be as important as the loss curve itself. A loss curve shows what happened to the objective; governance telemetry helps explain how the training system responded while that objective evolved.

The results also suggest a broader architectural implication. Future LLM training stacks may benefit from separating the optimizer plane from the governance plane. The optimizer plane computes parameter updates. The governance plane monitors training state, interprets instability, and applies bounded corrective action. This separation allows existing optimizers such as AdamW to remain in place while adding a control layer that improves runtime behavior. Such a design is especially relevant when training becomes more expensive, longer, and more operationally fragile. In that setting, the goal is not only to optimize a loss function, but also to preserve productive compute and reduce the probability of failed or degraded runs.

This paper has several important limitations. First, the experiments are primarily controlled single-GPU stress tests. They are useful for isolating behavior, but they do not fully represent distributed frontier-scale pretraining environments. Multi-GPU and multi-node settings introduce additional sources of instability, including communication overhead, synchronization issues, hardware faults, data pipeline bottlenecks, and checkpoint-recovery complexity. LBW-Guard may be relevant to these settings, but this paper does not yet validate that broader claim.

Second, most experiments use LoRA-based training. Although the no-LoRA TinyLlama 1B sanity check suggests that the method is not structurally dependent on adapters, the evidence remains limited. A stronger validation would include additional full-parameter training runs, more model families, longer training horizons, and multiple datasets. The current evidence supports a scoped claim about stability-sensitive LLM training behavior, not a universal claim about all training regimes.

Third, the paper specifies LBW-Guard at the component-control level rather than disclosing the full proprietary controller implementation. This is a deliberate boundary between scientific specification and product-level intellectual property protection. The paper discloses the architecture, component roles, telemetry categories, bounded-control framing, public run settings, result tables, and reproducibility artifact. However, it does not provide full code-level reproducibility of the internal controller policy. Readers should therefore interpret the reproducibility claim as partial rather than complete.

Fourth, the gradient-clipping comparison is informative but not exhaustive. The paper tests clipping at selected values under the main stress condition, but it does not perform a fully tuned clipping sweep across many thresholds, schedules, models, and learning rates. It is therefore possible that a carefully tuned clipping recipe could improve some AdamW baselines. The present claim is narrower: ordinary global clipping, as tested here, does not reproduce LBW-Guard's trainability preservation under stress.

Fifth, several experimental settings are intentionally stress-oriented and should not be treated as recommended default training recipes. The aggressive learning-rate conditions are designed to expose differences in failure behavior, not to propose that practitioners should normally train with those settings. This stress-test design is appropriate for evaluating training-control governance, because governance is most valuable when the training process approaches instability. Nevertheless, the distinction between stress testing and recommended training practice should remain clear.

Sixth, the statistical evidence remains preliminary. The paper includes seed repeatability evidence for the prior 3B setting, but broader statistical validation is still needed. Future work should include more seeds, confidence intervals, additional datasets, longer runs, and independent reproduction. This is particularly important because training instability can be sensitive to initialization, data order, hardware behavior, and implementation details.

\section{Conclusion}

This paper introduced LBW-Guard as a bounded autonomous training-control governance layer for LLM training under stress. The central contribution is both architectural and empirical: AdamW remains responsible for parameter updates, while LBW-Guard governs the runtime conditions under which those updates are executed. This separation reframes training stability as a systems problem, not merely an optimizer-selection problem.

Across controlled stress tests centered on Qwen2.5-7B and extended to 3B and 14B model-size settings, LBW-Guard improves final perplexity across all tested model scales. Under learning-rate stress, it preserves trainability in regimes where AdamW becomes severely degraded. The gradient-clipping comparison suggests that the effect is not reducible to ordinary local gradient suppression, and the no-LoRA TinyLlama-1B full-parameter sanity check suggests that the behavior is not purely adapter-specific.

The empirical contribution is therefore not simply that LBW-Guard obtains lower perplexity in selected runs. More importantly, it shows that bounded control over optimizer execution can change the runtime behavior of training itself. Instead of treating instability as noise, failure, or an after-the-fact debugging problem, LBW-Guard treats instability as a training-state signal that can be sensed, interpreted, and governed during execution. The question is not only whether an optimizer can compute effective updates, but whether the training process can remain productive when the trajectory becomes stressed, oscillatory, or fragile.

This framing also changes what should be measured. Conventional optimizer evaluation usually emphasizes final loss, perplexity, throughput, and sometimes memory footprint. A training-control governance method should additionally report control-active steps, regime switches, bounded scale behavior, and other telemetry that makes the control process observable. Such telemetry does not replace standard quality metrics, but it helps explain how the training system behaves while learning unfolds. In this paper, the reported control-active behavior supports the interpretation that LBW-Guard is not merely a passive wrapper around AdamW, but an active governance layer over training execution.

The broader implication is that future training stacks may need to evolve beyond optimizer-only design. As models become larger, training runs become longer, and compute becomes more expensive, the cost of instability grows. Training systems may therefore require a layered architecture in which the optimizer remains responsible for parameter updates, while a separate governance plane monitors the run, detects fragile operating conditions, and applies bounded corrective action. In this view, the optimizer is not replaced; it is governed.

The evidence remains deliberately scoped. The present experiments are controlled single-GPU stress tests, mostly using LoRA-based training, with one no-LoRA sanity check. They do not prove universal superiority across all optimizers, architectures, datasets, hardware environments, or frontier-scale distributed pretraining regimes. Within the tested settings, however, the results support the view that bounded training-control governance can improve stability-sensitive LLM training, preserve productive compute under stress, and remain distinct from optimizer replacement.

Future work should evaluate LBW-Guard across broader settings, including additional model families, full-parameter training, multi-GPU and multi-node environments, longer training horizons, additional datasets, and more extensive seed-based statistical validation. Further work should also compare LBW-Guard against stronger tuned stabilization baselines, alternative optimizers, dynamic learning-rate strategies, and production-scale recovery mechanisms. These extensions would help determine whether training-control governance can become a general architectural layer for reliable large-scale model training.

Overall, LBW-Guard provides an initial empirical case for Learn-by-Wire training: a training paradigm in which the learning process is not only optimized, but also sensed, interpreted, and governed. As LLM development becomes increasingly costly and operationally fragile, training-control governance may become an essential complement to optimization itself.

\section*{Broader Impact}

This work develops training-control infrastructure for improving the stability, observability, and productive compute efficiency of large language model training. The positive impacts are primarily operational and environmental. By reducing failed or severely degraded runs, bounded training-control governance may help reduce wasted accelerator time, lower experimentation cost, and improve the reliability of model-development workflows. Better training observability may also help researchers and engineers detect instability earlier and intervene in more disciplined ways.

The potential negative impacts are indirect. More efficient and more reliable training infrastructure may reduce the cost of developing capable models, including models that could be misused. Improved training efficiency may also accelerate competitive pressure to train larger models, which could partially offset compute-efficiency gains. This paper does not release a new pretrained model, dataset, or application-facing system. The evaluated contribution is a training-control method, and the empirical claims are scoped to controlled stress-test settings rather than deployment of a public model.

Overall, the broader impact of LBW-Guard depends on how training-control governance is used. When applied responsibly, it may improve resource efficiency, reliability, and engineering transparency. When applied without governance at the organizational level, the same efficiency gains could contribute to faster development of models whose downstream risks require separate evaluation and oversight.

\section*{Assets, Licenses, and Compute Resources}
Experiments use publicly available model and dataset assets, including Qwen2.5 model variants, TinyLlama, WikiText-103, CUDA, PyTorch AdamW, and LoRA. Users should comply with the corresponding model, dataset, and software licenses. The \lbw{} implementation evaluated in this paper is proprietary and is not released with the submission; the paper reports the component-level method specification, public controller configuration, and experimental results needed to interpret the claims. Experiments were run on a single CUDA GPU environment; wall-clock time values are reported in the result tables.

A public Colab test script is archived on Zenodo to support inspection of the experimental workflow and execution pathway \citep{radianis2026zenodo}. The artifact complements the paper's component-level method specification and reported result tables. It does not disclose the full proprietary LBW-Guard controller implementation, and therefore the reproducibility level should be understood as partial rather than complete code-level reproducibility.

\bibliographystyle{unsrtnat}
\bibliography{references}

\appendix

\section{Base Run Reference Settings}
\label{app:settings}

Table~\ref{tab:settings} reports the main reference settings used across the controlled stress-and-robustness experiments. These settings are provided to support reproducibility and to clarify the experimental boundary of the reported results. The table should be interpreted as the public experimental configuration for the paper, while the proprietary LBW-Guard controller policy is not fully disclosed.

\begin{table}[h]
\centering
\small
\begin{tabular}{p{0.48\linewidth}p{0.40\linewidth}}
\toprule
\textbf{Category} & \textbf{Setting / value} \\
\midrule
Primary LR/clip model & Qwen/Qwen2.5-7B \\
Model-size sweep & Qwen2.5-3B, Qwen2.5-7B, Qwen2.5-14B \\
Device & CUDA \\
Dataset & WikiText-103 raw \\
Train/Eval data & Full train split / full validation split \\
Validation & WikiText validation perplexity \\
Steps & 1000 \\
Schedule & all cosine \\
Primary 7B LR suite sequence length & 128 \\
Primary 7B LR suite batch / gradient accumulation & 2 / 2 \\
LoRA reference & $r=16$, alpha=64, dropout=0.05 \\
Gradient clipping baselines & AdamW $g=1.0$ and $g=0.5$ \\
\bottomrule
\end{tabular}
\caption{Main experimental reference settings.}
\label{tab:settings}
\end{table}

\section{Public Execution Interface}
\label{app:execution_interface}

The following code illustrates the public optimizer-construction interface used to compare AdamW and LBW-Guard. The purpose of this interface is to show that the AdamW configuration remains unchanged across the baseline and governed variants. LBW-Guard receives the same trainable parameters, learning rate, AdamW momentum coefficients, and weight decay, while adding bounded control-governance arguments for telemetry and runtime modulation.

This code is provided as an interface-level reproducibility example rather than a disclosure of the full proprietary LBW-Guard controller policy.

\begin{verbatim}
# AdamW baseline
optimizer = torch.optim.AdamW(
    params,
    lr=LR,
    betas=(BETA1, BETA2),
    weight_decay=WEIGHT_DECAY,
)

# LBW-Guard
optimizer = lbw.Guard(
    params,
    lr=LR,
    betas=(BETA1, BETA2),
    weight_decay=WEIGHT_DECAY,
    auto_enabled=True,
    stats_freq=LBW_STATS_FREQ,
    stress_threshold=LBW_STRESS_TH,
    spike_threshold=LBW_SPIKE_TH,
    recovery_fast=LBW_REC_FAST,
    ema_decay=LBW_EMA_DECAY,
    use_max_rms=True,
)


\end{verbatim}

The AdamW baseline uses the standard PyTorch optimizer interface. LBW-Guard preserves the same optimizer-facing hyperparameters and adds control-specific configuration for sensing frequency, stress/spike detection, recovery behavior, exponential moving average smoothing, and telemetry. These arguments define the public control interface, not the full internal controller implementation.

\section{Reproducibility Artifact}
\label{app:reproducibility}

A lightweight Colab-based reproduction artifact is archived on Zenodo \citep{radianis2026zenodo} at URL https://doi.org/10.5281/zenodo.20174991. The artifact is provided to support external inspection of the experimental workflow, configuration structure, evaluation pathway, and AdamW versus LBW-Guard comparison procedure. It is not the full internal experiment code used to produce every result in the paper.

The artifact may produce different numerical values from those reported in the paper because public Colab execution depends on available hardware, runtime environment, library versions, random seeds, and resource constraints. The purpose of the artifact is therefore to support partial reproducibility and methodological transparency, not exact replication of the original controlled run.

The proprietary LBW-Guard controller policy is not fully disclosed. The paper instead provides the public component-level specification, experimental settings, result tables, telemetry categories, and bounded-control interpretation required to assess the method.

\end{document}